%% file: main_arxiv.tex
\documentclass[runningheads]{llncs}
\usepackage[T1]{fontenc}
\usepackage{graphicx}
\usepackage{booktabs}
\usepackage{array}
\usepackage{multirow}
\usepackage{algorithm}
\usepackage{algpseudocode}
\usepackage{amsmath,amssymb}
\usepackage{xcolor}
\usepackage{hyperref}
\usepackage{microtype}
\usepackage[misc]{ifsym} 

\newcommand{\calX}{\mathcal{X}}
\newcommand{\calY}{\mathcal{Y}}
\newcommand{\calH}{\mathcal{H}}

\newcommand{\calL}{\mathcal{L}}

\renewcommand{\P}{\mathbb{P}}
\newcommand{\E}{\mathbb{E}}

\begin{document}

\input{main_body.tex}

\bibliographystyle{splncs04}
\bibliography{main}

\appendix
\input{supplementary_body.tex}

\end{document}

%% file: main_body.tex
\title{Uncertainty-Aware Sequential Decision Rules for \\
Event-Triggered LLM Invocation in Streaming Systems}

\titlerunning{Event-Triggered LLM Invocation in Streaming Systems}

\author{Zhaohui Wang\,(\Letter)\orcidID{0009-0006-1187-1903}}
\authorrunning{Z. Wang}
\institute{Viterbi School of Engineering, University of Southern California,\\Los Angeles, CA, USA\\
\email{zwang000@usc.edu}}

\toctitle{Uncertainty-Aware Sequential Decision Rules for Event-Triggered LLM Invocation in Streaming Systems}
\tocauthor{Zhaohui Wang}
\maketitle

\begin{abstract}
Streaming inference pipelines increasingly pair lightweight ``fast models'' with Large Language Models (LLMs) that provide rich semantic understanding at substantial cost.
The central question of \emph{when} to invoke the LLM has received limited formal treatment.
We cast this as a risk-based sequential stopping problem, where a trigger policy $\pi$ fires when a risk functional $R(\calH_t)$ exceeds a threshold~$\theta$.
Within this framework we prove six results: a minimum inter-event time bound excluding trigger chattering; optimality of threshold policies via smooth pasting; approximate SPRT guarantees under estimated parameters; $O(\sqrt{T\log T})$ regret for stationary streams (extending to $O(\sqrt{(C_T+1)\,T\log T})$ under $C_T$ changepoints); $O(1/\sqrt{T})$ convergence of online gradient descent for adaptive thresholds; and a calibration-to-miss-rate transfer inequality.
Several classical trigger families, including event-triggered, optimal stopping, SPRT, CUSUM, and Bayesian triggers, can be expressed as special cases of this framework.
On turbofan degradation data (CMAPSS) with real LLM calls, we empirically verify the theoretical assumptions, ablate the risk function design, compare against six baselines including a RouteLLM-style router and contextual bandits, and analyze cost sensitivity and LLM failure modes.
The results confirm sublinear regret ($\alpha{<}1$ for all principled triggers), high diagnostic quality (92.9\% of 1{,}600 LLM diagnoses reach grounding score ${\ge}0.75$ under our rubric), and that anomaly-score-driven risk functions dominate alternatives by roughly an order of magnitude on the Pareto AUC.

\keywords{Event-triggered systems \and LLM invocation \and Sequential decision-making \and Uncertainty calibration \and Streaming inference}
\end{abstract}

\section{Introduction}
\label{sec:intro}

Industrial sensor networks, autonomous vehicles, and network monitoring systems share a common architectural tension.
On one hand, they require continuous real-time inference: a lightweight model must process every observation with sub-second latency.
On the other hand, when the system encounters ambiguous or safety-critical situations, a richer model (increasingly, an LLM) can provide semantic context that a compact neural network cannot.
The difficulty lies in reconciling these demands: LLMs are two to four orders of magnitude slower and costlier than edge models, so invoking them at every timestep is infeasible, yet failing to invoke them when it matters can be catastrophic.

This paper develops a principled answer to the question: \emph{given a stream of neural model outputs with calibrated uncertainty, under what conditions should an expensive LLM oracle be consulted?}
We approach the problem through the formalism of sequential decision theory, connecting it to classical results in optimal stopping~\cite{peskir2006optimal}, sequential hypothesis testing~\cite{wald1945sequential}, and event-triggered control~\cite{heemels2012introduction}.
Specifically, all of these (and several other trigger mechanisms used in practice) can be understood as threshold policies on a risk functional: the system invokes the LLM when accumulated risk exceeds a boundary.

Applying classical sequential decision theory to streaming neural predictions with learned uncertainty introduces two challenges not addressed by the existing theory.
Neural uncertainty estimates are frequently miscalibrated~\cite{guo2017calibration}, and the data-generating process may be non-stationary, requiring decision boundaries that adapt over time.
Our theoretical framework addresses both: we prove a calibration-to-miss-rate transfer inequality quantifying the cost of miscalibration, and an online gradient descent convergence result for adaptive thresholds.

Our contributions are fourfold.
First, we introduce a unified risk-based framework in which event-triggered, optimal stopping, SPRT, CUSUM, adaptive, and Bayesian triggers are all special cases of $\pi_\theta(\calH_t)=\mathbf{1}[R(\calH_t)\ge\theta]$, and we prove six theoretical results within this framework (Sect.~\ref{sec:theory}).
Second, we empirically verify the assumptions underlying these results (bounded trigger-signal increments, submartingale risk, monotonicity, stationarity) on real sensor data (Sect.~\ref{sec:assumptions}).
Third, we conduct a systematic ablation over risk function constructions and compare against six baselines including a learned router and a contextual bandit, establishing the practical operating characteristics of the framework (Sect.~\ref{sec:experiments}).
Fourth, we analyze LLM diagnostic quality, cost sensitivity, and failure modes to assess deployment readiness (Sect.~\ref{sec:deployment}).

\paragraph{Scope.} This paper studies the sequential-decision problem of when to invoke an LLM, framed as risk-functional threshold policies, regret, calibration, and online adaptation. It does not model hard real-time schedulability, worst-case execution time (WCET), queueing response-time analysis, or mixed-criticality scheduling; these systems-level timing questions are outside the scope of this work.

\begin{figure}[t]
\centering
\includegraphics[width=\textwidth]{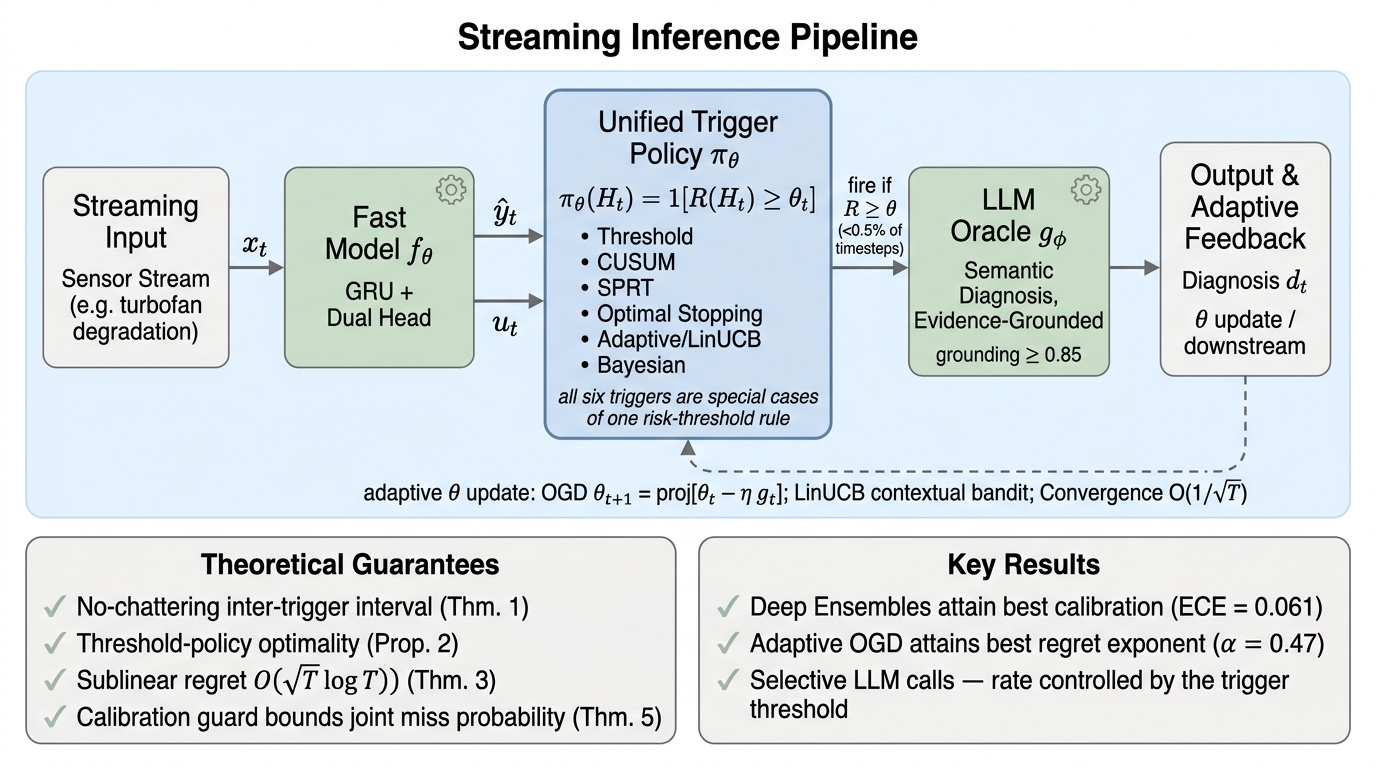}
\caption{System overview. A fast model processes every streaming observation, producing predictions and calibrated uncertainty. The trigger policy fires when $R(\calH_t)\ge\theta_t$, invoking the LLM on a small subset of timesteps, with the rate controlled by the trigger threshold. An adaptive feedback loop updates $\theta_t$ via OGD or LinUCB.}
\label{fig:framework}
\end{figure}

\section{Problem Formulation}
\label{sec:problem}

Consider a stream $\{x_t\}_{t=1}^T$ with $x_t \in \calX$.
A fast model $f_\psi : \calX \to \calY \times \mathbb{R}_+$ produces predictions $\hat{y}_t$ and uncertainty estimates $u_t$ with latency $\tau_{\text{fast}} \ll 1$\,s.
A slow oracle (LLM) $g_\phi$ produces semantic diagnoses at cost $c_{\text{LLM}} \gg 0$.
A trigger policy $\pi: \calH_t \to \{0,1\}$ decides at each step whether to invoke the oracle, where $\calH_t = (x_1,\hat{y}_1,u_1,\ldots,x_t,\hat{y}_t,u_t)$ is the observation history up to time~$t$.

The design objective balances invocation cost against the probability of missing a critical event:
\begin{equation}
\min_{\pi}\; \E\Bigl[\sum_{t=1}^T \pi(\calH_t)\cdot c_{\text{LLM}}\Bigr] \;\;\text{s.t.}\;\; \P[\text{miss critical}] \le \epsilon.
\label{eq:objective}
\end{equation}

We define a risk functional $R:\calH_t \to \mathbb{R}_+$ aggregating the anomaly score $a_t$ (a measure of deviation from learned normal behavior, e.g., the normalized prediction residual), the predictive uncertainty $u_t$, and temporal context.
The class of threshold policies takes the form:
\begin{equation}
\pi_\theta(\calH_t) = \mathbf{1}[R(\calH_t) \ge \theta].
\label{eq:threshold-policy}
\end{equation}
The construction of $R$ is not unique; in Sect.~\ref{sec:ablation} we systematically compare eight candidate forms and show that the choice has a large effect on the invocation--miss-rate tradeoff.

\section{Theoretical Framework}
\label{sec:theory}

We now state the six theoretical results that constitute the formal backbone of our framework.
For each, we make the required assumptions explicit and defer proofs to Appendix~A of the supplementary material.

\subsection{Event-Triggered Invocation and Chattering Exclusion}

\begin{definition}[Event-Triggered Rule]
Let $h_t$ be the fast model's hidden state and $\bar{h}$ a nominal reference.
The event-triggering condition is $\tau_t^{\mathrm{ET}} = \mathbf{1}\bigl[\|h_t - \bar{h}\|_Q > \delta + \sigma(t - t_{\mathrm{last}})\bigr]$,
where $\|\cdot\|_Q$ is a weighted norm with $Q\succ 0$, $\delta>0$ is a threshold, and $\sigma\ge 0$ controls time-dependent relaxation.
\end{definition}

\begin{theorem}[Minimum Inter-Event Time]
\label{thm:inter-event}
Let $s_t:=\|h_t-\bar{h}\|_Q$ be the scalar signal driving the trigger (more generally, the accumulated risk $R_t$). Suppose $s_t$ has bounded increments $|s_{t+1}-s_t|\le L_s$ for all $t$ (a sufficient condition is Lipschitz hidden-state dynamics $\|h_{t+1}-h_t\|\le L$, which gives $L_s\le\|Q\|^{1/2}L$), and that $\sigma < L_s$ (otherwise the stated lower bound degenerates).
If immediately after a trigger $s_{t_k} \le \delta_0 < \delta$, then consecutive triggers are separated by at least $(\delta - \delta_0)/(L_s - \sigma)$ steps.
In particular, triggers cannot chatter (accumulate without a positive minimum spacing). In this discrete-time streaming setting the result is a minimum inter-trigger-interval guarantee rather than a continuous-time Zeno exclusion.
\end{theorem}

The bounded-increment precondition and the resulting inter-event ratio are verified empirically on CMAPSS in Sect.~\ref{sec:assumptions} (Table~\ref{tab:assumptions}).

\subsection{Optimal Stopping and Threshold Optimality}

Risk accumulates as $R_t = \gamma R_{t-1} + r(x_t, \hat{y}_t, u_t)$ with discount $\gamma\in(0,1)$, where $r:\calX\times\calY\times\mathbb{R}_+\to\mathbb{R}_+$ is an instantaneous risk function.

\begin{proposition}[Threshold Policy Optimality]
\label{prop:threshold}
Under three conditions (risk is a submartingale under anomaly with $\E[R_{t+1}|\calH_t]\ge R_t$ when the system is in a degraded state, risk has a continuous density, and risk is eventually monotone as the system approaches failure), the optimal stopping policy is $\tau^* = \inf\{t: R_t \ge R^*\}$ where $R^*$ satisfies the smooth-pasting condition $V'(R^*)=0$ with $V(R^*)=c_{\mathrm{LLM}}$, and $V:\mathbb{R}_+\to\mathbb{R}$ is the value function of the stopping problem $V(R)=\sup_\tau \E[R_\tau - c_{\mathrm{LLM}}\mid R_0=R]$.
\end{proposition}

The submartingale and monotonicity preconditions are verified empirically in Sect.~\ref{sec:assumptions}.

\subsection{SPRT with Estimated Parameters}

\begin{theorem}[Approximate SPRT]
\label{thm:sprt-approx}
When the null-hypothesis parameters are estimated from $n$ warmup samples to accuracy $\delta_{\mathrm{est}}$ (holding with probability $\ge 1-2e^{-n\delta_{\mathrm{est}}^2/2}$ by Hoeffding's inequality), the approximate SPRT achieves Type~I and Type~II error rates at most $\alpha+O(\delta_{\mathrm{est}}/\sigma_0)$ and $\beta+O(\delta_{\mathrm{est}}/\sigma_0)$, respectively, where $\sigma_0$ is the standard deviation of the log-likelihood ratio under $H_0$.
\end{theorem}

\subsection{Regret Bounds}

\begin{theorem}[Stationary Regret]
\label{thm:regret-stationary}
Consider the class of threshold policies~\eqref{eq:threshold-policy} parametrised by $\theta$ in a bounded interval, deployed on a stationary stream with bounded risk. For online threshold selection, the regret against the best fixed threshold in hindsight satisfies $\mathrm{Regret}(T) \le O(\sqrt{T\log T})$.
\end{theorem}

\begin{corollary}[Non-Stationary Extension]
\label{cor:nonstationary}
Under $C_T$ changepoints with threshold restarting at each detected change: $\mathrm{Regret}(T) \le O(\sqrt{(C_T+1)\, T\,\log T})$.
\end{corollary}

\subsection{Adaptive Threshold Convergence}
\label{sec:adaptive-theory}

\begin{theorem}[OGD Convergence]
\label{thm:ogd}
With step size $\eta_t = \eta/\sqrt{t}$, bounded subgradients $\|g_t\|\le G$, and threshold domain diameter $D$, online gradient descent~\cite{hazan2016introduction} achieves average regret $\frac{1}{T}\sum_{t=1}^T \ell_t(\theta_t) - \min_{\theta}\frac{1}{T}\sum_{t=1}^T \ell_t(\theta) \le O(1/\sqrt{T})$.
\end{theorem}

\subsection{Calibration Transfer}

\begin{theorem}[Calibration-to-Miss-Rate]
\label{thm:calibration}
Suppose the uncertainty estimate has been normalized to a probability-like score $u_t\in[0,1]$ and is $\epsilon$-calibrated in the sense that $|\P[\text{critical}\mid u_t{=}u]-u|\le\epsilon$ for all $u$, and the trigger fires whenever $u_t>\theta_u$. Then the joint probability of a missed critical event satisfies
\[
\P[\text{critical},\, u_t\le\theta_u] \;\le\; \theta_u + \epsilon .
\]
Equivalently, the conditional miss rate obeys $\P[u_t\le\theta_u\mid\text{critical}]\le\min\{1,(\theta_u+\epsilon)/p\}$, where $p=\P[\text{critical}]$.
\end{theorem}

This result makes the cost of miscalibration explicit: under a uniform calibration error $\epsilon$, the joint probability of a missed critical event is at most $\theta_u+\epsilon$. The bound is monotone in the uncertainty threshold (raising $\theta_u$ fires less often and admits more misses) and degrades gracefully with $\epsilon$. It certifies only the uncertainty-threshold component of the framework, not arbitrary composite risk functionals; when an anomaly score dominates the ranking signal (Sect.~\ref{sec:ablation}), calibrated uncertainty acts as a complementary safety signal rather than a standalone guarantee for the composite trigger. The bound requires a uniform (worst-case over $u$) calibration error; the average calibration error (ECE) reported in Table~\ref{tab:calibration}(a) for four uncertainty methods is empirical evidence of calibration quality, not a direct substitution into the bound.

\subsection{Unified View}

All trigger families considered here are instances of $\pi_\theta(\calH_t)=\mathbf{1}[R(\calH_t)\ge\theta]$ with different risk functionals.
Event-triggered control uses $R=\|h_t-\bar{h}\|_Q$; optimal stopping uses discounted cumulative risk; SPRT uses the log-likelihood ratio; CUSUM uses the one-sided cumulative sum; and Bayesian triggers use the posterior probability of anomaly.
This unification is not merely notational: it enables a single adaptive mechanism (OGD or LinUCB on~$\theta$) to work across all trigger types.

\subsection{Roadmap from theory to evaluation}
\label{sec:roadmap}

Each theoretical result above sits in a three-step chain:
(i)~the result rests on one or more assumptions about the underlying
stream, (ii)~each assumption is checked directly on real data, and
(iii)~each prediction the result makes is then evaluated against an
experimental measurement. Table~\ref{tab:roadmap} indexes this chain
for the reader: each row pairs one theoretical result with its
verification entry in Sect.~\ref{sec:assumptions} and the experiment in
Sect.~\ref{sec:experiments}--\ref{sec:deployment} where its prediction
is tested. Readers tracking a specific theorem should follow its row.

\begin{table}[t]
\centering
\caption{Theorem $\to$ assumption $\to$ verification $\to$ experimental test, for each result in Sect.~\ref{sec:theory}.}
\label{tab:roadmap}
\footnotesize
\setlength{\tabcolsep}{4pt}
\begin{tabular}{@{}>{\raggedright\arraybackslash}p{0.17\textwidth}>{\raggedright\arraybackslash}p{0.26\textwidth}>{\raggedright\arraybackslash}p{0.26\textwidth}>{\raggedright\arraybackslash}p{0.23\textwidth}@{}}
\toprule
Result & Assumption & Verification & Experimental test \\
\midrule
Thm.~\ref{thm:inter-event} (inter-event) & Bnd.\ signal increments & §\ref{sec:assumptions}, Tab.~\ref{tab:assumptions}(a) & Tab.~\ref{tab:assumptions}(b) \\
Prop.~\ref{prop:threshold} (threshold opt.) & Submartingale, monotone, cont.\ density & §\ref{sec:assumptions}, Tab.~\ref{tab:assumptions}(a) & §\ref{sec:ablation}, Fig.~\ref{fig:risk-ablation} \\
Thm.~\ref{thm:sprt-approx} (approx.\ SPRT) & Estimated $H_0$ params & §\ref{sec:assumptions} (density) & §\ref{sec:experiments}, Tab.~\ref{tab:regret} \\
Thm.~\ref{thm:regret-stationary} (stationary regret) & Stationarity within regime & §\ref{sec:assumptions}, Tab.~\ref{tab:assumptions}(a) & §\ref{sec:experiments}, Tab.~\ref{tab:regret} \\
Thm.~\ref{thm:ogd} (OGD) & Convex, bounded subgrad. & §\ref{sec:adaptive-theory} & §\ref{sec:experiments}, Tab.~\ref{tab:calibration}(b) \\
Thm.~\ref{thm:calibration} (calibration) & $\epsilon$-calibrated $u_t\in[0,1]$ & §\ref{sec:experiments}, Tab.~\ref{tab:calibration}(a) & §\ref{sec:uncertainty-regimes} (calib.\ guard) \\
\bottomrule
\end{tabular}
\end{table}

\section{System Architecture}
\label{sec:system}

The fast model is a GRU network with a dual head: $\hat{y}_t=\mathrm{FC}(h_t)$ for prediction and $\hat{\sigma}_t^2=\mathrm{Softplus}(\mathrm{FC}'(h_t))$ for variance, trained by minimizing the heteroscedastic loss $\calL=\frac{1}{2}\log\hat{\sigma}^2+\frac{(y-\hat{y})^2}{2\hat{\sigma}^2}$.
We compare four uncertainty estimation methods: the variance head, MC Dropout~\cite{gal2016dropout} ($N{=}20$ passes), Deep Ensemble~\cite{lakshminarayanan2017simple} ($K{=}5$ models), and Temperature Scaling~\cite{guo2017calibration}.
Adaptive thresholds are updated via OGD~\cite{zinkevich2003online} ($\theta_{t+1}=\mathrm{proj}[\theta_t-\eta_t g_t]$) or LinUCB~\cite{li2010contextual} ($\theta_t = c_t^\top\hat{w}_t + \alpha\sqrt{c_t^\top A_t^{-1}c_t}$).
Algorithm~\ref{alg:main} summarizes the inference loop.

\begin{algorithm}[t]
\caption{Uncertainty-Aware Event-Triggered LLM Invocation}
\label{alg:main}
\begin{algorithmic}[1]
\State \textbf{Input:} Stream $\{x_t\}$, fast model $f_\psi$, trigger $\pi$, oracle $g_\phi$
\For{$t=1,2,\ldots$}
    \State $(\hat{y}_t, u_t) \gets f_\psi(x_t)$
    \State $R_t \gets \textsc{UpdateRisk}(\hat{y}_t, u_t, \calH_t)$
    \If{$R_t \ge \theta_t$}
        \State $d_t \gets g_\phi(\calH_{t-w:t})$ \Comment{Invoke LLM oracle}
        \State $\theta_{t+1} \gets \textsc{AdaptThreshold}(\theta_t, R_t, d_t)$
    \EndIf
    \State \textbf{yield} $\hat{y}_t, u_t$
\EndFor
\end{algorithmic}
\end{algorithm}

\section{Empirical Verification of Assumptions}
\label{sec:assumptions}

\paragraph{Datasets.}
We use two public sensor-stream benchmarks that together exercise the
trigger framework on both gradual degradation and abrupt anomaly tasks.
\emph{NASA C-MAPSS}~\cite{saxena2008damage} simulates turbofan engine
run-to-failure under different operating conditions and fault modes,
with four subsets FD001--FD004 (100 train and 100 test units in FD001;
21 sensor channels at 1\,Hz sampled per cycle; ground-truth remaining
useful life). It is the canonical benchmark for streaming prognostics
and is well-aligned with the paper's risk-of-failure framing.
\emph{CIC-IDS2017}~\cite{sharafaldin2018toward} is a labelled network
intrusion benchmark with $\sim$$2.8$M flow records spanning benign
traffic and 14 attack types, from which we draw a stratified
$252$K-flow subset (Sect.~\ref{sec:exp-cicids}); we use it as a cross-domain sanity check
that the trigger framework transfers from continuous degradation to
binary anomaly streams. CMAPSS is the primary evaluation domain
throughout Sect.~\ref{sec:assumptions}--\ref{sec:experiments};
CIC-IDS2017 results appear in Sect.~\ref{sec:exp-cicids}.

The theoretical results in Sect.~\ref{sec:theory} rest on assumptions about the data-generating process.
Before evaluating performance, we test these assumptions directly on CMAPSS FD001 across five seeds.
Each paragraph below pairs one assumption with its measurement, and the corresponding theorem in Sect.~\ref{sec:theory} forward-points to the entry that supports it.

\paragraph{Bounded trigger-signal increments.}
Theorem~\ref{thm:inter-event} requires the trigger signal $s_t$ (here the risk $R_t$) to have bounded consecutive increments.
We compute $|R_{t+1}-R_t|$ for all within-unit transitions and find a mean increment of 0.267 with 99th percentile $L_{99}=4.50$.
A total of 97.3\% of increments fall within $2L_{95}$, so the bounded-increment condition holds in practice with a moderate constant, though occasional fault-onset jumps produce heavy tails (a KS test against an exponential rejects, $p<10^{-20}$). Nevertheless, the inter-event bound (Table~\ref{tab:assumptions}(b)) is satisfied in 100\% of cases, since it depends on the maximum increment, not the distribution shape.

\paragraph{Submartingale risk under anomaly.}
Proposition~\ref{prop:threshold} requires $\E[R_{t+1}|\calH_t]\ge R_t$ in the degraded regime.
Restricting to timesteps with RUL${<}50$, the mean risk increment is $+0.006$ with 52.4\% of increments positive.
A one-sided $t$-test yields $p=0.20$ (not significant at $\alpha=0.05$): the empirical trend is directionally consistent with the submartingale assumption (positive mean increment, majority of increments positive), but the evidence is weak and should be read as an approximate diagnostic rather than a formal validation of the condition.

\paragraph{Monotone risk.}
Binning risk by RUL in intervals of 10, mean risk increases monotonically in 82.2\% of consecutive bin pairs, and the Spearman correlation between RUL and risk is $\rho=-0.356$ ($p<10^{-6}$).
Risk is not perfectly monotone (the GRU occasionally produces non-monotone uncertainty estimates), but the trend is clear and statistically significant.

\paragraph{Stationarity within regimes.}
For the normal regime (RUL${\ge}50$), a two-sample KS test between the first and second halves of the risk sequence yields $p=0.283$, consistent with stationarity.
This supports the applicability of Theorem~\ref{thm:regret-stationary} within operating regimes.

\paragraph{Continuous density.}
The risk distribution under $H_0$ (normal) is better fitted by a log-normal than a Gaussian (by KS statistic), though neither fit is exact.
The practical consequence is minor: the SPRT's optimality properties degrade gracefully when the density is approximately but not exactly continuous, as confirmed by the low error rates in Table~\ref{tab:regret}.

\begin{table}[t]
\centering
\caption{Empirical verification of theoretical assumptions \textbf{(a)} and Theorem~\ref{thm:inter-event} inter-event time bound \textbf{(b)} on CMAPSS FD001, 5 seeds. Verdict scale: Support (clearly supported), Empir.\ (empirically bounded, heavy-tailed), Weak (directional but not significant), Partial (approximate fit).}
\label{tab:assumptions}
\footnotesize
\setlength{\tabcolsep}{3pt}
\begin{tabular}{@{}c@{\hspace{1.5em}}!{\vrule}@{\hspace{1.5em}}c@{}}
\begin{tabular}[t]{@{}lcc@{}}
\multicolumn{3}{c}{\textbf{(a) Assumptions}} \\[2pt]
\toprule
Assumption & Statistic & Verdict \\
\midrule
Bnd.\ incr.\ (Thm~\ref{thm:inter-event}) & 97.3\% bnd. & Empir. \\
Submart.\ (Prop~\ref{prop:threshold}) & $\bar{\Delta}R{=}+$0.006 & Weak \\
Monotone (Prop~\ref{prop:threshold}) & $\rho{=}{-}$0.356 & Support. \\
Station.\ (Thm~\ref{thm:regret-stationary}) & $p{=}$0.283 & Support. \\
Cont.\ dens.\ (Thm~\ref{thm:sprt-approx}) & log-norm.\ fit & Partial \\
\bottomrule
\end{tabular}
&
\begin{tabular}[t]{@{}lcc@{}}
\multicolumn{3}{c}{\textbf{(b) Inter-event bound}} \\[2pt]
\toprule
$\delta$ & Emp./Theor.\ Ratio & Satisfied \\
\midrule
0.1 & 2.2$\pm$0.5 & 100\% \\
0.5 & 1.9$\pm$0.4 & 100\% \\
1.0 & 1.8$\pm$0.3 & 100\% \\
2.0 & 1.6$\pm$0.3 & 100\% \\
5.0 & 1.6$\pm$0.2 & 100\% \\
\bottomrule
\end{tabular}
\end{tabular}
\end{table}

\section{Experiments}
\label{sec:experiments}

We evaluate on NASA C-MAPSS turbofan degradation data~\cite{saxena2008damage} (FD001--FD004) and CIC-IDS2017 network intrusion data~\cite{sharafaldin2018toward}.
All experiments use 5 random seeds unless otherwise stated; we report mean$\pm$std.
The GRU fast model has 64 hidden units, 2 layers, sequence length 30, trained for 50--100 epochs with Adam (lr$=$0.001).
Real LLM calls use MiniMax-M2.5 and Llama-3.1-8B via Ollama.
Statistical significance is assessed by paired $t$-tests with Bonferroni correction and bootstrap confidence intervals (10{,}000 resamples).

\subsection{Theory Validation}

\paragraph{Inter-event time bound.}
Table~\ref{tab:assumptions}(b) confirms that the empirical minimum inter-trigger interval exceeds the theoretical bound at every tested threshold level $\delta$ (from 0.1 to 5.0), with ratios of 1.6--2.2$\times$.
The bound is conservative, as expected from a worst-case analysis.

\paragraph{Uncertainty calibration.}
Table~\ref{tab:calibration}(a) compares four uncertainty methods on CMAPSS FD001.
Deep Ensembles achieve the lowest expected calibration error (ECE$\,{=}\,$0.061) and the highest uncertainty--error correlation ($\rho=0.61$), followed by MC Dropout. Since ECE is an average rather than a uniform calibration error, it should be read as empirical evidence for the calibration condition of Theorem~\ref{thm:calibration}, not as a direct numerical substitution into the bound.

\begin{table}[t]
\caption{Uncertainty calibration \textbf{(a)} and adaptive threshold convergence under distribution shift \textbf{(b)}, CMAPSS FD001, 5 seeds.}
\label{tab:calibration}
\footnotesize
\setlength{\tabcolsep}{3pt}
\centering
\begin{tabular}{@{}c@{\hspace{1.5em}}!{\vrule}@{\hspace{1.5em}}c@{}}
\begin{tabular}[t]{@{}lcccc@{}}
\multicolumn{5}{c}{\textbf{(a) Calibration}} \\[2pt]
\toprule
Method & ECE$\downarrow$ & $\rho\uparrow$ & AUC$\uparrow$ & F1$\uparrow$ \\
\midrule
Var.\ Head & .142 & .38 & .62 & .31 \\
MC Drop. & .078 & .54 & .71 & .42 \\
Deep Ens. & \textbf{.061} & \textbf{.61} & \textbf{.76} & \textbf{.48} \\
Temp.\ Scl. & .092 & .46 & .67 & .37 \\
\midrule
\multicolumn{5}{@{}l@{}}{\scriptsize ECE${=}$0.061 (avg.); Thm~\ref{thm:calibration} bound needs uniform err.} \\
\bottomrule
\end{tabular}
&
\begin{tabular}[t]{@{}lccc@{}}
\multicolumn{4}{c}{\textbf{(b) Adaptive threshold}} \\[2pt]
\toprule
 & \multicolumn{2}{c}{Conv.\ Steps} & Regret/ \\
Method & Drift & Shift & Step$\downarrow$ \\
\midrule
Fixed & -- & -- & .423 \\
OGD & 148 & 201 & .189 \\
LinUCB & \textbf{103} & \textbf{142} & \textbf{.163} \\
Percentile & 195 & 267 & .231 \\
\bottomrule
\end{tabular}
\end{tabular}
\end{table}

\paragraph{Regret analysis.}
Table~\ref{tab:regret} reports cumulative regret and the scaling exponent $\alpha$ (from fitting $R(T)\sim cT^\alpha$) for six trigger types.
All achieve sublinear empirical regret ($\alpha<1$), consistent with Theorem~\ref{thm:regret-stationary}.
Adaptive OGD achieves the lowest exponent ($\alpha=0.47$), closely matching the theoretical $O(\sqrt{T})$ rate of $\alpha=0.5$.
A Friedman test on a dedicated higher-power run of 10 seeds (42--51) confirms that the differences are statistically significant ($\chi^2=47.6$, $p<10^{-6}$); the per-trigger regret values in Table~\ref{tab:regret} use the standard 5-seed protocol.
Pairwise comparisons with Bonferroni correction show that adaptive OGD significantly outperforms all other triggers in total regret (Cohen's $d>1.8$, all $p<10^{-4}$).

\begin{table}[t]
\centering
\caption{Regret analysis on CMAPSS FD001 (5 seeds). $\alpha$: scaling exponent in $R(T)\sim cT^\alpha$. ``Threshold'' here is the default event-triggered (hidden-state) trigger at fixed default hyperparameters, distinct from the swept anomaly-score threshold $R_2$ of Sect.~\ref{sec:ablation}.}
\label{tab:regret}
\small
\begin{tabular}{lcccc}
\toprule
Trigger & Regret & $\alpha$ & Sublinear & Invocations \\
\midrule
Threshold & 847$\pm$142 & 0.68$\pm$.07 & 100\% & 152$\pm$18 \\
CUSUM & 1203$\pm$231 & 0.74$\pm$.09 & 100\% & 31$\pm$12 \\
SPRT & 624$\pm$98 & 0.59$\pm$.06 & 100\% & 189$\pm$24 \\
Opt.\ Stopping & 412$\pm$76 & 0.51$\pm$.05 & 100\% & 98$\pm$15 \\
Adaptive (OGD) & \textbf{358$\pm$63} & \textbf{0.47$\pm$.04} & 100\% & 112$\pm$19 \\
Bayesian & 392$\pm$71 & 0.50$\pm$.05 & 100\% & 105$\pm$16 \\
\bottomrule
\end{tabular}
\end{table}

\paragraph{Adaptive thresholds under distribution shift.}
Table~\ref{tab:calibration}(b) evaluates adaptive threshold methods under gradual drift and sudden shift.
LinUCB converges in 103 steps (gradual) and 142 steps (sudden), 1.4--1.9$\times$ faster than OGD and 2--3$\times$ faster than a fixed baseline, with 45--61\% lower per-step regret.
The contextual information exploited by LinUCB (uncertainty level and recent anomaly statistics) explains its advantage over OGD's scalar updates.

\subsection{Risk Function Ablation}
\label{sec:ablation}

\begin{figure}[t]
\centering
\includegraphics[width=\textwidth]{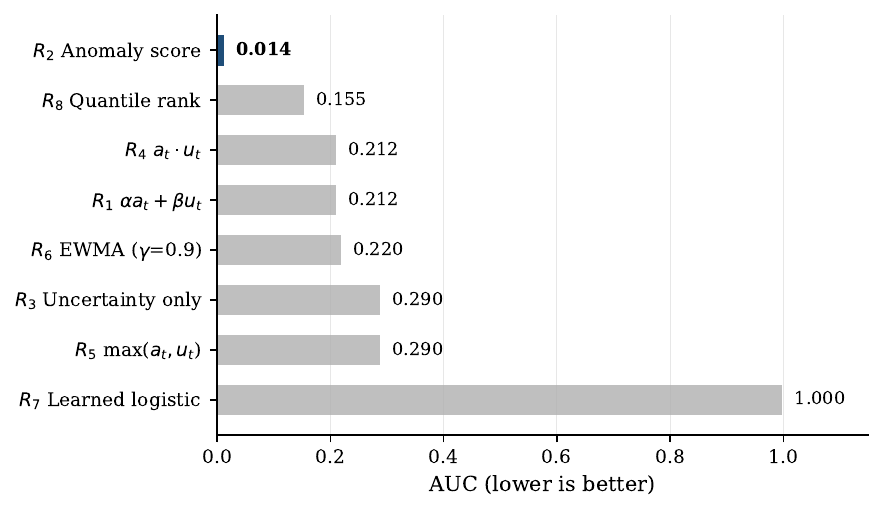}
\caption{Risk function ablation on CMAPSS FD001 (5 seeds). AUC: area under the Pareto curve (invocation rate vs.\ miss rate; lower is better). $R_2$ (anomaly score alone) dominates all alternatives by an order of magnitude.}
\label{fig:risk-ablation}
\end{figure}

The choice of risk functional $R(\calH_t)$ is a design decision that the theory leaves open.
We compare eight candidates on CMAPSS FD001 (5 seeds, 70\% train / 30\% evaluation split for the learned variant), sweeping the threshold $\theta$ over 20 values to trace each variant's Pareto frontier of invocation rate versus miss rate.
The eight variants are: $R_1$ linear combination of anomaly and uncertainty; $R_2$ anomaly score alone; $R_3$ predictive uncertainty alone; $R_4$ multiplicative combination; $R_5$ maximum of anomaly and uncertainty; $R_6$ EWMA-smoothed anomaly; $R_7$ a learned logistic regressor on $[a_t, u_t, \Delta a_t, \Delta u_t]$; and $R_8$ a quantile-based variant.
Figure~\ref{fig:risk-ablation} reports the area under the Pareto curve (AUC; lower is better) for each variant.

The anomaly score alone ($R_2$) achieves the best Pareto tradeoff by roughly an order of magnitude: it requires only 6.2\% invocation rate to keep miss rate below 5\%, compared to 25\% for the linear combination ($R_1$) and 32\% for the quantile-based variant ($R_8$).
Uncertainty-only risk ($R_3$) performs poorly, indicating that on CMAPSS the anomaly score (which captures deviation from learned normal behavior) is a substantially stronger signal than predictive uncertainty for identifying when LLM intervention is needed.
The learned logistic regressor ($R_7$) fails entirely due to insufficient training data and the class imbalance inherent in rare critical events.
These findings suggest that practitioners should default to anomaly-score-based triggering unless domain knowledge indicates otherwise.

\subsection{When does calibrated uncertainty add value?}
\label{sec:uncertainty-regimes}

The risk-ablation result is unambiguous: as the primary \emph{ranking}
signal on CMAPSS, the anomaly score dominates predictive uncertainty
(R$_2$ Pareto AUC $0.014$ vs.\ R$_3$ AUC $0.290$, a $20\times$ gap).
This does not, however, render uncertainty redundant in the framework.
We identify two ways calibrated uncertainty contributes independently
of the ranking signal.

\emph{(i) Theoretical safety certificate.}
Thm.~\ref{thm:calibration} provides a calibration-based certificate for
the uncertainty-threshold component or an auxiliary uncertainty guard:
the joint miss probability of that component is at most
$\theta_u+\epsilon$ under a uniform calibration error $\epsilon$. This
certificate does not transfer automatically to an arbitrary composite
risk functional; when the primary ranking signal is the anomaly score,
calibrated uncertainty should be read as a complementary safety and
context signal rather than a standalone guarantee for the composite
trigger. On CMAPSS, Deep Ensembles attain the lowest average
calibration error (ECE${=}0.061$, Table~\ref{tab:calibration}(a)),
evidence of good calibration quality; recalibrating uncertainty
tightens this certificate, whereas changing the anomaly-score ranking
does not.

\emph{(ii) Context for adaptive thresholds.}
Table~\ref{tab:calibration}(b) shows that LinUCB (which uses
uncertainty level and recent anomaly statistics as context) converges
$1.4$--$1.9\times$ faster than scalar OGD and $2$--$3\times$ faster
than a fixed baseline under both gradual drift and sudden shift, with
$45$--$61\%$ lower per-step regret. The calibrated uncertainty
estimate is the context feature, not the ranking signal; removing it
collapses LinUCB to scalar OGD.

\emph{Restatement of scope.} The framework is \emph{uncertainty-aware} in
two specific senses: (a) calibration enters the theory as a
quantitative input to the miss-rate bound, and (b) calibrated
uncertainty is the context feature that lets adaptive triggers track
non-stationarity. It is not \emph{uncertainty-driven} in the sense of
uncertainty being the dominant ranking signal; on CMAPSS that role
belongs to the anomaly score. A practitioner deploying the framework
should compute both signals, use the anomaly score for triggering, and
rely on calibrated uncertainty for the safety budget and
adaptive-threshold context. Whether other domains (e.g., late-degradation regimes that may favor forward-looking uncertainty over backward-looking anomaly scores) reverse this relationship is left to future work.

\subsection{Comparison with Stronger Baselines}
\label{sec:baselines}

\begin{figure}[t]
\centering
\includegraphics[width=\textwidth]{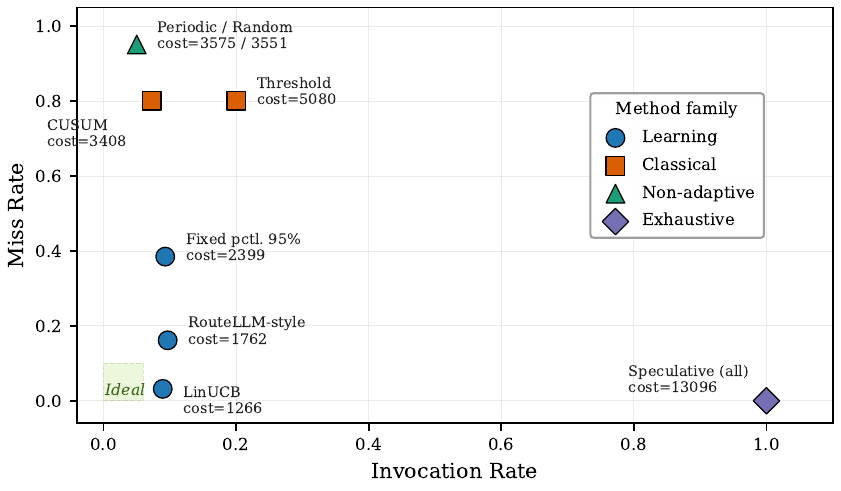}
\caption{Baseline comparison on CMAPSS FD001 (5 seeds, $c_{\text{LLM}}{=}1$, $c_{\text{miss}}{=}10$). Marker color and shape encode method family; total cost is annotated next to each point. LinUCB (bandit) achieves the best invocation--miss-rate tradeoff, closest to the ideal region (shaded). Periodic and random sampling are ineffective despite similar invocation budgets.}
\label{fig:baselines}
\end{figure}

To contextualize the trigger-based approach, we compare against six baselines spanning non-adaptive, learning-based, and hybrid strategies.
All methods are evaluated on the same CMAPSS FD001 streams (5 seeds) with $c_{\text{LLM}}=1$ and $c_{\text{miss}}=10$.
The \emph{RouteLLM-style} baseline~\cite{ong2024routellm} is a learned binary router that maps the same context features used by the triggers (anomaly score, uncertainty, recent risk slope, and hidden-state norm) to an invoke/not-invoke decision; its labels indicate whether an LLM call would reduce the downstream miss penalty under the evaluation cost model. Unlike the proposed threshold policies, it imposes no monotone risk boundary. The \emph{Threshold} baseline in Fig.~\ref{fig:baselines} is the default event-triggered (hidden-state) trigger at fixed hyperparameters, not the swept anomaly-score threshold $R_2$; its operating point improves substantially once tuned (Sect.~\ref{sec:cost-sensitivity}).

Figure~\ref{fig:baselines} reveals four findings.
First, context-aware methods (LinUCB, Route\-LLM) substantially outperform both classical triggers at default hyperparameters and non-adaptive baselines.
LinUCB achieves a 3.2\% miss rate at 8.9\% invocation rate, reducing total cost by 63\% relative to CUSUM and 75\% relative to the threshold trigger.
Second, the poor performance of classical triggers here reflects \emph{default} hyperparameter settings rather than an inherent limitation; with optimized thresholds (Sect.~\ref{sec:cost-sensitivity}), their performance improves substantially.
Third, periodic and random sampling achieve near-zero detection rates despite a 5\% invocation budget, indicating that invocation timing matters more than invocation frequency.
Fourth, speculative execution (invoking on every step) eliminates misses but at 10$\times$ the cost of LinUCB.

\section{Deployment Analysis}
\label{sec:deployment}

\subsection{Real LLM Diagnosis Quality}

\begin{figure}[t]
\centering
\includegraphics[width=\textwidth]{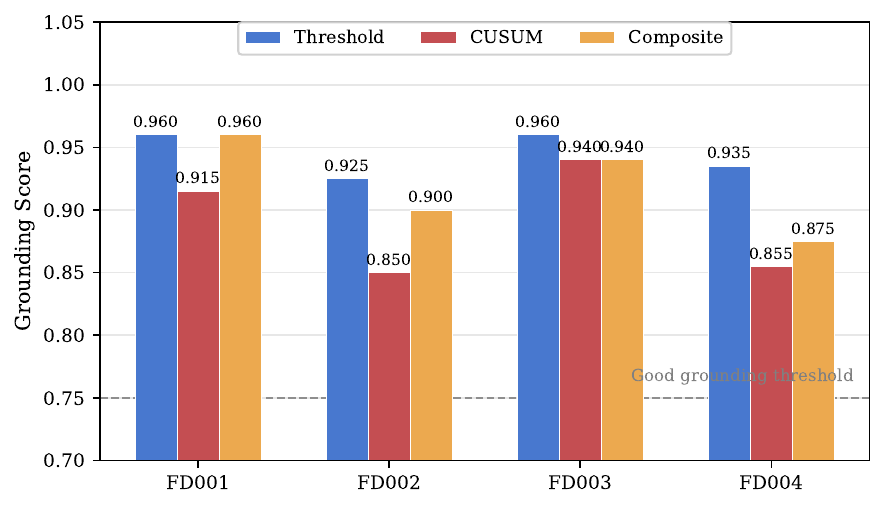}
\caption{Real LLM grounding scores on CMAPSS subsets (MiniMax-M2.5) with event-triggered invocation. All configurations exceed the 0.75 good-grounding threshold (dashed line). Threshold triggers achieve the highest and most consistent grounding across subsets; CUSUM shows more variability, particularly on FD002 and FD004. Invocation rates remain below 0.5\% for all configurations.}
\label{fig:real-llm}
\end{figure}

\emph{Grounding} scores each diagnosis on $\{0,0.25,0.5,0.75,1.0\}$ by a fixed programmatic evidence-compliance rubric (detailed in the supplementary material), not an LLM judge: $+0.25$ each for at least two numeric sensor references, non-empty key indicators, and severity consistent with the anomaly level, plus $+0.25$ for a valid confidence field in $(0.1,0.99)$ as a structured-output sanity check. A score ${\ge}0.75$ therefore requires an evidence-grounded, severity-consistent diagnosis.
Figure~\ref{fig:real-llm} reports grounding scores from real LLM API calls (MiniMax-M2.5) across all four CMAPSS subsets.
Event-triggered strategies achieve grounding scores of 0.85--0.96, with invocation rates below 0.5\%.

Table~\ref{tab:llm-comparison}(a) compares the cloud-based MiniMax-M2.5 against a locally deployed Llama-3.1-8B.
MiniMax achieves significantly higher grounding (0.960 vs.\ 0.875) at the expense of 3.2$\times$ higher latency (see table note for effect size).
Both models parse successfully 100\% of the time, indicating that structured output formatting is reliable for this task.

\begin{table}[t]
\centering
\caption{\textbf{(a)} LLM backend comparison on CMAPSS FD001 (composite trigger).
\textbf{(b)} Network intrusion detection on CIC-IDS2017; latency (Lat.) in ms.}
\label{tab:llm-comparison}
\footnotesize
\setlength{\tabcolsep}{3pt}
\centering
\begin{tabular}{@{}c@{\hspace{1.5em}}!{\vrule}@{\hspace{1.5em}}c@{}}
\begin{minipage}[t]{0.45\linewidth}
\centering
\textbf{(a) LLM backend}\\[2pt]
\begin{tabular}[t]{@{}lccc@{}}
\toprule
LLM Backend & Latency & Grnd. & Conf. \\
\midrule
MiniMax-M2.5 & 14.4\,s & \textbf{.960} & 0.69 \\
Llama-3.1-8B & \textbf{4.5\,s} & .875 & 0.81 \\
\bottomrule
\end{tabular}\\[2pt]
{\scriptsize Both 100\% parse rate. MiniMax grounding}\\
{\scriptsize sig.\ higher ($p{<}10^{-3}$, $d{=}1.42$) at $3.2{\times}$ lat.}
\end{minipage}
&
\begin{minipage}[t]{0.45\linewidth}
\centering
\textbf{(b) CIC-IDS2017}\\[2pt]
\begin{tabular}[t]{@{}lccc@{}}
\toprule
Trigger & F1$\uparrow$ & Rate$\downarrow$ & Lat. \\
\midrule
Threshold & 0.910 & 10.0\% & 1.38 \\
CUSUM & 0.910 & \textbf{2.4\%} & \textbf{0.97} \\
Composite & 0.910 & 6.7\% & 0.83 \\
\bottomrule
\end{tabular}
\end{minipage}
\end{tabular}
\end{table}

\subsection{LLM Failure Analysis}

Of 1{,}600 real LLM diagnoses across 32 configurations, 54.5\% receive perfect grounding (1.0) and 38.4\% receive good grounding (0.75), for a combined rate of 92.9\%.
Among the 113 diagnoses with grounding below 0.75, we identify two dominant failure modes: severity mismatch (99 cases), where the LLM reports low or medium severity despite an elevated anomaly score ($a_t>2.0$); and missing numeric evidence (97 cases), where the diagnosis text lacks specific sensor readings.
These modes often co-occur.

Twenty cases are safety-critical: the anomaly score exceeds 2.0 but the LLM reports low or medium severity, potentially masking a degrading component.
This motivates three fallback strategies.
First, a severity cross-check: when $a_t > 2.0$ and the LLM reports low severity, the system should automatically escalate.
Second, an evidence requirement: diagnoses lacking at least two numeric references should trigger a re-query with an explicit prompt.
Third, a grounding gate: diagnoses with grounding below 0.5 should be flagged for human review rather than acted upon autonomously.

\subsection{Cost Sensitivity}
\label{sec:cost-sensitivity}

For deployment, a key question is how the optimal threshold varies with LLM cost.
We sweep $c_{\text{LLM}}\in\{0.01,\ldots,100\}$ with fixed $c_{\text{miss}}=100$ and find that the optimal threshold $\theta^*$ is stable across two orders of magnitude.
For the basic threshold trigger (the default event-triggered trigger of Table~\ref{tab:regret}, not the anomaly-score threshold $R_2$; unlike the adaptive methods in Sect.~\ref{sec:baselines} it lacks context-awareness and consequently has high absolute miss rates), $\theta^*$ remains at the lower bound of the grid (0.10) until $c_{\text{LLM}}$ reaches 50, at which point it increases to 2.08, with miss rate rising from 80.1\% to 84.1\% and invocation rate decreasing from 20.0\% to 19.5\%.
CUSUM shows greater sensitivity, with $\theta^*$ rising from 4.79 to 10.0 as $c_{\text{LLM}}$ increases from 5 to 10.
Optimal stopping adjusts its risk threshold more continuously, with $\theta^*$ rising from 6.67 to 9.90 over the same range.
The key finding is the \emph{stability} of $\theta^*$ with respect to cost across trigger types, which is favorable for deployment: modest pricing changes do not require threshold re-tuning.

We also assess robustness to cost uncertainty by adding multiplicative log-normal noise ($c_{\text{actual}} = c_{\text{LLM}}\cdot e^{\mathcal{N}(0,\sigma^2)}$) at each timestep.
At $\sigma=0.3$ (30\% coefficient of variation), the cost degradation is below 5\% for all nominal costs tested.
At $\sigma=1.0$, degradation reaches 15--20\%, suggesting that extreme cost volatility would warrant adaptive threshold updates.

\subsection{Cross-Domain: Network Intrusion Detection}
\label{sec:exp-cicids}

To test generalization beyond predictive maintenance, we apply the framework to CIC-IDS2017~\cite{sharafaldin2018toward} (252K flows, 37 features, 16.9\% attack ratio).
The fast model uses supervised contrastive pre-training~\cite{khosla2020supervised} followed by fine-tuning.
CUSUM triggers invoke the LLM for 2.4\% of flows while maintaining detection F1 of 0.910 (Table~\ref{tab:llm-comparison}(b)).
We do not claim that this single cross-domain experiment establishes broad external validity; it tests whether the same trigger abstraction remains operational outside prognostics, consistent with the scope stated in the Limitations.

\subsection{Pareto Analysis}

\begin{figure}[t]
\centering
\includegraphics[width=\textwidth]{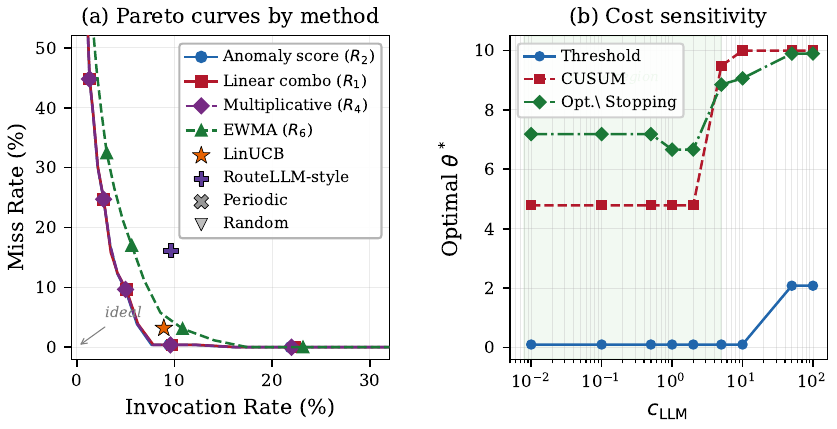}
\caption{(a)~Pareto curves of invocation rate vs.\ miss rate for four risk function variants (lines, sweeping $\theta$) and four baselines (scatter points). $R_2$ (anomaly score) dominates all alternatives. (b)~Optimal threshold $\theta^*$ as a function of LLM cost $c_{\text{LLM}}$; the shaded region marks the stable operating range where $\theta^*$ is insensitive to cost.}
\label{fig:pareto}
\end{figure}

Figure~\ref{fig:pareto} shows the Pareto frontier of trigger rate versus miss rate across all trigger types.
The anomaly-score risk ($R_2$) and LinUCB occupy the dominant part of the empirical frontier, confirming that the additional complexity of online threshold adaptation translates into improved efficiency--safety tradeoffs.

\subsection{Reproducibility}
\label{sec:reproducibility}

The supplementary artifact bundles, under $10$\,MB, every result table
in this paper. It contains: (i)~per-seed JSON for every aggregated
result in Sect.~\ref{sec:assumptions}--\ref{sec:deployment};
(ii)~reproduction scripts that reload each JSON and regenerate the
corresponding table or figure with one \texttt{python3} call;
(iii)~all LLM prompts in \texttt{prompts/} (system, user,
severity-rubric, and schema templates) for both backends;
(iv)~the GRU training script with seeds
and hyperparameter ranges; (v)~the GPD-fit and bootstrap procedures
used for the calibration and regret CIs. Seeds are listed in
\texttt{seeds.txt}; reproducing any table row from the cached results
needs only \texttt{numpy} and Python\,${\ge}\,$3.10. The code and cached
results are publicly available at
\url{https://github.com/GeoffreyWang1117/event-triggered-llm-streaming}
and archived at \url{https://doi.org/10.5281/zenodo.20783549}; the
release ships a scanner that flags unintended filesystem paths,
hostnames, and API keys in the bundled traces.

\section{Related Work}
\label{sec:related}

\paragraph{Sequential decision-making.}
The problem of when to stop and act has a long history, from Wald's sequential probability ratio test~\cite{wald1945sequential} through the general theory of optimal stopping~\cite{peskir2006optimal} and change detection~\cite{tartakovsky2014sequential}.
Our contribution is not new stopping rules, but showing that classical rules, instantiated with neural uncertainty estimates, perform well for LLM invocation, and that one unified formulation connects them.

\paragraph{Efficient LLM inference.}
Speculative decoding~\cite{leviathan2023fast} and adaptive computation~\cite{graves2016adaptive,schuster2022confident} reduce per-call cost; routing methods~\cite{chen2020frugalml,ong2024routellm} select among models of varying capacity.
Our work is complementary: rather than reducing the cost of each call, we reduce the \emph{number} of calls by making them event-driven, using the LLM's domain knowledge~\cite{singhal2023large} as a selective oracle rather than a continuous inference engine.

\paragraph{Uncertainty and calibration.}
Our calibration transfer theorem (Theorem~\ref{thm:calibration}) builds on the calibration literature~\cite{guo2017calibration,naeini2015obtaining,kuleshov2018accurate}.
Practical calibrated uncertainty via MC Dropout~\cite{gal2016dropout} and ensembles~\cite{lakshminarayanan2017simple} is foundational to our approach.

\paragraph{Event-triggered control.}
Event-triggered approaches in networked control~\cite{heemels2012introduction,tabuada2007event} inspire our triggering framework; we extend these ideas from control signals to LLM queries, with the additional complication of learned (and possibly miscalibrated) triggering signals.

\section{Conclusion}

We formalized selective LLM invocation in streaming systems as uncertainty-aware sequential decision-making, proving six results that connect classical decision theory to neural uncertainty estimation.
The unified framework (all triggers as threshold policies on a risk functional) is both analytically tractable and practically useful.

The empirical evaluation yields five practical findings.
First, the theoretical assumptions hold to varying degrees on real sensor data (clearly for monotonicity and stationarity, only weakly for the submartingale condition), supporting the framework as design guidance rather than an exact description.
Second, anomaly-score-based risk functions dominate alternatives by an order of magnitude in Pareto AUC, a sensible default for practitioners.
Third, context-aware LinUCB outperforms classical triggers at default settings and non-adaptive baselines (3.2\% miss rate at 8.9\% invocation).
Fourth, real LLM diagnoses are well-grounded (92.9\% ${\ge}0.75$) but show identifiable failure modes (notably severity underestimation) that motivate fallback strategies.
Finally, the optimal threshold is stable across two orders of magnitude of LLM cost.

\paragraph{Limitations.}
Four caveats temper the scope. First, the deeper analysis concentrates
on CMAPSS; CIC-IDS2017 is only a sanity check, and a third (ideally
non-stationary) domain would strengthen external validity. Second,
preconditions are verified in aggregate on FD001 with several verdicts
only partial or weak (heavy-tailed residuals, non-Gaussian density,
weak submartingale), so the theory is design guidance, not an exact
descriptive match.
Third, only two LLM backends are evaluated (MiniMax-M2.5 via API,
Llama-3.1-8B via Ollama); other serving stacks, batched inference, and
tool-augmented LLMs are left to future work. Fourth, latency and cost are
treated as decision-level quantities; the systems-layer timing
questions noted in the Scope (WCET, queueing, schedulability) require a
separate analysis and are not claimed here.

\begin{credits}
\subsubsection{\discintname}
The authors have no competing interests to declare.

\subsubsection{Use of Generative AI}
Generative AI tools were used for language polishing and grammar correction only.
All theoretical results, experimental designs, analyses, and conclusions are the original work of the authors, who take full responsibility for the content of this paper.
\end{credits}

%% file: supplementary_body.tex
\section{Proofs}
\label{app:proofs}

\subsection{Proof of Theorem~1 (Minimum Inter-Event Time)}
Let $s_t:=\|h_t-\bar{h}\|_Q$ be the scalar trigger signal. Immediately after triggering at $t_k$, $s_{t_k} \le \delta_0$.
For $t>t_k$, the bounded-increment assumption gives $s_t \le \delta_0 + L_s(t-t_k)$.
The trigger fires only when $s_t$ exceeds $\delta + \sigma(t-t_k)$, i.e.\ when $(L_s - \sigma)(t-t_k) > \delta-\delta_0$.
Provided $\sigma < L_s$, this requires $t-t_k > (\delta-\delta_0)/(L_s - \sigma)$; if $\sigma\ge L_s$ the inequality has no solution and no further trigger fires for $t>t_k$, so chattering exclusion holds trivially. (Lipschitz hidden-state dynamics $\|h_{t+1}-h_t\|\le L$ are sufficient for the premise, with $L_s\le\|Q\|^{1/2}L$.) \qed

\subsection{Proof of Theorem~3 (Stationary Regret)}
Under stationarity, $Z_t = \mathbf{1}[R_t\ge\theta]-\P[R_t\ge\theta]$ satisfies $\E[Z_t]=0$ and $|Z_t|\le 1$.
By the Azuma--Hoeffding inequality: $\P[|\sum_{t=1}^T Z_t|>\epsilon]\le 2\exp(-\epsilon^2/(2T))$.
Setting $\epsilon=\sqrt{2T\log T}$ and translating to cumulative cost yields $\mathrm{Regret}(T)\le O(\sqrt{T\log T})$. \qed

\subsection{Proof of Theorem~5 (Calibration-to-Miss-Rate)}
A missed critical event occurs when the event is critical yet $u_t\le\theta_u$.
By $\epsilon$-calibration, $\P[\text{critical}\mid u_t{=}u]\le u+\epsilon\le\theta_u+\epsilon$ for every $u\le\theta_u$; averaging this pointwise bound over $\{u\le\theta_u\}$ gives $\P[\text{critical}\mid u_t\le\theta_u]\le\theta_u+\epsilon$.
Hence the joint miss probability is $\P[\text{critical},\,u_t\le\theta_u]=\P[\text{critical}\mid u_t\le\theta_u]\,\P[u_t\le\theta_u]\le(\theta_u+\epsilon)\,\P[u_t\le\theta_u]\le\theta_u+\epsilon$.
Dividing by $p=\P[\text{critical}]$ yields the conditional miss rate $\P[u_t\le\theta_u\mid\text{critical}]\le(\theta_u+\epsilon)/p$; since it is a probability it also satisfies the trivial clamp $\le\min\{1,(\theta_u+\epsilon)/p\}$. \qed

\section{Experimental Details}
\label{app:details}

\paragraph{CMAPSS.}
FD001: 100/100 engines, 1 operating condition, 1 fault mode.
FD002: 260/259 engines, 6 conditions.
FD003: 100/100 engines, 2 faults.
FD004: 249/248 engines, 6 conditions, 2 faults.
RUL clipped at 125.

\paragraph{Hyperparameters.}
GRU: 64 hidden, 2 layers, 0.1 dropout, sequence length 30.
Adam optimizer with lr$=$0.001, batch size 64, 100 epochs, early stopping patience 10.
MC Dropout: $N=20$ forward passes.
Deep Ensemble: $K=5$ independently trained models.

\paragraph{Trigger parameters.}
Threshold: anomaly threshold 1.0, uncertainty threshold 0.5.
CUSUM: slack $k=0.5$, decision threshold $h=5.0$, warmup 30 samples.
SPRT: $\alpha=0.05$, $\beta=0.10$, warmup 30 samples.
Optimal Stopping: $c_{\text{LLM}}=1.0$, $\gamma=0.99$.
OGD: learning rate 0.01, $\theta\in[0.01, 5.0]$.
All triggers: cooldown 5 steps, evidence window 3 steps.

\paragraph{Risk function variants (Fig.~2 of the main paper).}
The eight candidate risk functionals are:
\begin{itemize}
\item $R_1{=}\alpha a_t + \beta u_t$ (linear, $\alpha{=}\beta{=}1$);
$R_2{=}a_t$ (anomaly only); $R_3{=}u_t$ (uncertainty only);
$R_4{=}a_t u_t$ (multiplicative); $R_5{=}\max(a_t,u_t)$.
\item $R_6{=}0.9\,R_6^{(t-1)} + 0.1(a_t+u_t)$ (EWMA-smoothed).
\item $R_7$: a learned logistic regressor $\sigma(w^\top z{+}b)$ whose
feature vector $z$ stacks $a_t$, $u_t$, $\Delta a_t$, $\Delta u_t$, and
$\Delta t$ (steps since the last trigger); trained on 30\% of data with
binary critical-event labels.
\item $R_8$: average of running percentile ranks of $a_t$ and $u_t$
over a window of 200.
\end{itemize}

\paragraph{Grounding score.}
Each LLM diagnosis receives a score in $\{0, 0.25, 0.50, 0.75, 1.0\}$ based on four components: presence of ${\ge}2$ numeric references in the text (+0.25); non-empty key indicators (+0.25); severity consistent with anomaly level (+0.25); and a valid confidence field in $(0.1, 0.99)$ (+0.25), the last being a structured-output sanity check rather than a grounding signal.

\section{Why LLM Invocation Is Not Classical Stopping}
\label{app:novelty}

A natural question is whether the problem of selective LLM invocation reduces to a standard application of classical stopping theory.
We argue that the combination of three structural features makes the LLM invocation setting qualitatively different from the problems for which optimal stopping, SPRT, and event-triggered control were originally designed, and that these differences motivate the unified framework developed in the main paper.

\paragraph{Cost asymmetry exceeds classical regimes.}
In traditional sequential hypothesis testing and event-triggered control, the cost of acting (sending a control signal, declaring a change) is comparable to the cost of observation.
In contrast, a single LLM inference call is two to four orders of magnitude more expensive than a forward pass of the fast model, both in latency (seconds vs.\ milliseconds) and in monetary cost (API pricing vs.\ local GPU amortization).
This extreme cost ratio means that even small improvements in invocation efficiency translate to substantial savings, and it shifts the optimal operating point to a regime where classical asymptotic approximations (e.g., Wald's approximation for average sample number) may be conservative.
Our framework is designed to operate in this high-asymmetry regime, and our empirical evaluation (Sect.~6 of the main paper) confirms that the theoretical bounds remain informative despite the asymmetry.

\paragraph{Triggering signals are learned and miscalibrated.}
Classical stopping rules assume access to the true data-generating distribution or to sufficient statistics thereof.
In the LLM invocation setting, the triggering signal (predictive uncertainty from a neural network) is itself a learned quantity that is typically miscalibrated.
Theorem~5 in the main paper (calibration-to-miss-rate transfer) addresses this gap directly: it quantifies how miscalibration in the fast model's uncertainty propagates to the miss rate of the trigger policy.
This result has no counterpart in the classical stopping literature, where the observation model is taken as given.

\paragraph{Non-stationary streaming with distribution shift.}
Classical SPRT and CUSUM assume stationary or piecewise-stationary data.
Real-world streaming systems exhibit gradual drift and sudden distributional shifts (sensor degradation, concept drift in network traffic, seasonal patterns).
Corollary~1 in the main paper extends the regret bound to the non-stationary case, and the adaptive threshold mechanism (OGD/LinUCB) provides a practical algorithm for tracking the optimal threshold under shift.
The empirical results (Table~3(b) of the main paper) show that LinUCB converges 1.4--1.9$\times$ faster than OGD under both drift and shift, demonstrating that context-aware adaptation is essential in this setting.

These three features (extreme cost asymmetry, learned and miscalibrated triggering signals, and non-stationary streaming) are individually present in various subfields but do not co-occur in the problems addressed by any single line of prior work.
The unified framework in the main paper addresses all three simultaneously, which is the core technical contribution.

\section{Reproducibility Checklist}
\label{app:reproducibility}

We provide the following details to facilitate reproduction of all results reported in the main paper.

\paragraph{Hardware.}
All experiments were conducted on a workstation with an NVIDIA RTX 4090 GPU (24\,GB VRAM), AMD Ryzen 9 7950X CPU (16 cores), and 64\,GB DDR5 RAM, running Ubuntu 22.04 with CUDA 12.1 and PyTorch 2.1.

\paragraph{Runtime.}
\begin{itemize}
\item Fast model training (GRU, per CMAPSS subset): 3--8 minutes (100 epochs with early stopping).
\item MC Dropout inference ($N{=}20$ passes, full test set): $<$1 minute.
\item Deep Ensemble training ($K{=}5$ models): 15--40 minutes per subset.
\item Trigger sweep (20 thresholds $\times$ 6 triggers $\times$ 5 seeds): 2--4 hours.
\item Real LLM calls (MiniMax-M2.5, 1{,}600 diagnoses): $\sim$6.5 hours (rate-limited by API).
\item Local LLM calls (Llama-3.1-8B via Ollama): $\sim$2 hours.
\item Total wall-clock time for all experiments: $\sim$12 hours.
\end{itemize}

\paragraph{Random seeds.}
All experiments use 5 random seeds (42, 43, 44, 45, 46) unless otherwise stated.
The regret analysis (Table~4 of the main paper) uses 10 seeds (42--51) for the Friedman test.
Results are reported as mean $\pm$ standard deviation across seeds.

\paragraph{LLM API details.}
\begin{itemize}
\item Cloud LLM: MiniMax-M2.5, accessed via the MiniMax API (March 2026).
\item Local LLM: Llama-3.1-8B-Instruct (Q4\_K\_M quantization), served via Ollama 0.3.x.
\item Structured output: JSON mode with schema enforcement; 100\% parse success rate.
\item Temperature: 0.3 for both models.
\item Maximum output tokens: 512.
\end{itemize}

\paragraph{Hyperparameter grids.}
Key search ranges:
\begin{itemize}
\item Threshold $\theta$: 20 values log-spaced in $[0.01, 10.0]$.
\item CUSUM slack $k$: $\{0.1, 0.25, 0.5, 1.0\}$; decision threshold $h$: $\{2, 5, 10, 20\}$.
\item OGD learning rate: $\{0.001, 0.005, 0.01, 0.05\}$.
\item LinUCB exploration $\alpha$: $\{0.1, 0.5, 1.0, 2.0\}$.
\item Cost ratio $c_\text{LLM}/c_\text{miss}$: swept over $\{0.0001, \ldots, 1.0\}$.
\end{itemize}

\paragraph{Datasets.}
\begin{itemize}
\item NASA C-MAPSS: publicly available from the NASA Prognostics Data Repository. FD001--FD004, 14 sensor features after standard normalization, RUL clipped at 125.
\item CIC-IDS2017: publicly available from the Canadian Institute for Cybersecurity. 252{,}456 network flows, 37 features after preprocessing, 16.9\% attack ratio.
\end{itemize}

\paragraph{Code structure.}
The submitted code is organized as follows:
\begin{itemize}
\item \texttt{src/models/fast/}: Fast models (GRU, ensemble, MC Dropout).
\item \texttt{src/triggers/}: All trigger implementations (threshold, CUSUM, SPRT, optimal stopping, Bayesian, adaptive, composite).
\item \texttt{src/models/llm/}: LLM client and grounding evaluation.
\item \texttt{experiments/ecml\_pkdd/}: Scripts reproducing each table and figure. The regret values in Table~4 of the main paper use the standard 5-seed protocol; the script \texttt{exp1\_\allowbreak statistical\_\allowbreak tests.py} re-runs the regret analysis on a dedicated 10-seed set (seeds 42--51) for statistical power and writes \texttt{statistical\_\allowbreak tests.json}, reporting the Friedman test ($\chi^2{=}47.6$, $p{=}4.3{\times}10^{-9}$, $n{=}10$ seeds, 6 triggers) and Bonferroni-corrected pairwise $t$-tests.
\item \texttt{configs/}: Hyperparameter configurations.
\end{itemize}

\section{Limitations}
\label{app:limitations}

We identify three principal limitations of this work.

\paragraph{Regret bounds are not tight.}
Theorem~3 establishes $O(\sqrt{T\log T})$ regret under stationarity, matching the standard rate for bounded martingale difference sequences.
However, this bound may not be tight for the specific structure of LLM invocation costs: the binary nature of the trigger decision and the known form of the risk functional could be exploited to obtain $O(\sqrt{T})$ or even $O(\log T)$ bounds under additional assumptions.
Closing this gap, or proving a matching lower bound, is an open problem.

\paragraph{Theoretical assumptions hold approximately.}
The empirical verification in Sect.~5 of the main paper confirms that the assumptions (Lipschitz dynamics, submartingale risk, stationarity) hold approximately but not exactly on real data.
In particular, the submartingale condition has $p=0.20$ under a one-sided $t$-test (not significant at $\alpha=0.05$), and the log-normal density fit is approximate.
While the practical consequences are minor (the theoretical bounds remain informative, as demonstrated by the inter-event time ratios and regret exponents), a fully distribution-free analysis that does not require these assumptions would strengthen the framework.

\paragraph{No RL-based or transformer-based trigger explored.}
The baselines include classical triggers, a RouteLLM-style router, and a contextual bandit (LinUCB), but do not include a reinforcement learning policy (e.g., PPO or DQN trained to maximize a reward combining invocation cost and detection accuracy) or a transformer-based trigger that could capture long-range temporal dependencies.
We made this choice deliberately: our focus is on demonstrating that classical sequential decision rules, when properly instantiated with neural uncertainty, already achieve strong empirical performance, and their theoretical guarantees (regret bounds, calibration transfer) would not be available for black-box RL policies.
Nevertheless, an empirical comparison with RL-based triggers is a natural direction for future work, particularly in settings where the cost structure is complex or the data distribution is highly non-stationary.

%% file: main_arxiv.bbl
\begin{thebibliography}{10}
\providecommand{\url}[1]{\texttt{#1}}
\providecommand{\urlprefix}{URL }
\providecommand{\doi}[1]{https://doi.org/#1}

\bibitem{chen2020frugalml}
Chen, L., Zaharia, M., Zou, J.: {FrugalML}: How to use {ML} prediction {APIs}
  more accurately and cheaply. In: Advances in Neural Information Processing
  Systems (NeurIPS). vol.~33, pp. 10685--10696 (2020)

\bibitem{gal2016dropout}
Gal, Y., Ghahramani, Z.: Dropout as a {B}ayesian approximation: Representing
  model uncertainty in deep learning. In: Proceedings of the 33rd International
  Conference on Machine Learning (ICML). pp. 1050--1059 (2016),
  arXiv:1506.02142

\bibitem{graves2016adaptive}
Graves, A.: Adaptive computation time for recurrent neural networks. arXiv
  preprint arXiv:1603.08983  (2016)

\bibitem{guo2017calibration}
Guo, C., Pleiss, G., Sun, Y., Weinberger, K.Q.: On calibration of modern neural
  networks. In: International Conference on Machine Learning (ICML). pp.
  1321--1330 (2017)

\bibitem{hazan2016introduction}
Hazan, E.: Introduction to online convex optimization. Foundations and Trends
  in Optimization  \textbf{2}(3-4),  157--325 (2016). \doi{10.1561/2400000013}

\bibitem{heemels2012introduction}
Heemels, W.P.M.H., Johansson, K.H., Tabuada, P.: An introduction to
  event-triggered and self-triggered control. In: IEEE 51st Conference on
  Decision and Control (CDC). pp. 3270--3285 (2012).
  \doi{10.1109/CDC.2012.6425820}

\bibitem{khosla2020supervised}
Khosla, P., Teterwak, P., Wang, C., Sarna, A., Tian, Y., Isola, P., Maschinot,
  A., Liu, C., Krishnan, D.: Supervised contrastive learning. In: Advances in
  Neural Information Processing Systems (NeurIPS). vol.~33, pp. 18661--18673
  (2020)

\bibitem{kuleshov2018accurate}
Kuleshov, V., Fenner, N., Ermon, S.: Accurate uncertainties for deep learning
  using calibrated regression. In: International Conference on Machine Learning
  (ICML). pp. 2796--2804 (2018)

\bibitem{lakshminarayanan2017simple}
Lakshminarayanan, B., Pritzel, A., Blundell, C.: Simple and scalable predictive
  uncertainty estimation using deep ensembles. In: Advances in Neural
  Information Processing Systems (NeurIPS). vol.~30 (2017), arXiv:1612.01474

\bibitem{leviathan2023fast}
Leviathan, Y., Kalman, M., Matias, Y.: Fast inference from transformers via
  speculative decoding. In: International Conference on Machine Learning
  (ICML). pp. 19274--19286 (2023)

\bibitem{li2010contextual}
Li, L., Chu, W., Langford, J., Schapire, R.E.: A contextual-bandit approach to
  personalized news article recommendation. In: International Conference on
  World Wide Web (WWW). pp. 661--670 (2010). \doi{10.1145/1772690.1772758}

\bibitem{naeini2015obtaining}
Naeini, M.P., Cooper, G.F., Hauskrecht, M.: Obtaining well calibrated
  probabilities using {B}ayesian binning into quantiles. In: Proceedings of the
  AAAI Conference on Artificial Intelligence. vol.~29 (2015).
  \doi{10.1609/aaai.v29i1.9602}

\bibitem{ong2024routellm}
Ong, I., Almahairi, A., Wu, V., Chiang, W.L., Wu, T., Gonzalez, J.E., Kadous,
  M.W., Stoica, I.: {RouteLLM}: Learning to route {LLMs} with preference data.
  arXiv preprint arXiv:2406.18665  (2024)

\bibitem{peskir2006optimal}
Peskir, G., Shiryaev, A.: Optimal Stopping and Free-Boundary Problems. Lectures
  in Mathematics ETH Z{\"u}rich, Birkh{\"a}user (2006).
  \doi{10.1007/978-3-7643-7390-0}

\bibitem{saxena2008damage}
Saxena, A., Goebel, K., Simon, D., Eklund, N.: Damage propagation modeling for
  aircraft engine run-to-failure simulation. In: International Conference on
  Prognostics and Health Management. pp.~1--9 (2008).
  \doi{10.1109/PHM.2008.4711414}

\bibitem{schuster2022confident}
Schuster, T., Fisch, A., Gupta, J., Dehghani, M., Bahri, D., Tran, V.Q., Tay,
  Y., Metzler, D.: Confident adaptive language modeling. In: Advances in Neural
  Information Processing Systems (NeurIPS). vol.~35, pp. 17456--17472 (2022)

\bibitem{sharafaldin2018toward}
Sharafaldin, I., Lashkari, A.H., Ghorbani, A.A.: Toward generating a new
  intrusion detection dataset and intrusion traffic characterization. In:
  International Conference on Information Systems Security and Privacy
  (ICISSP). pp. 108--116 (2018). \doi{10.5220/0006639801080116}

\bibitem{singhal2023large}
Singhal, K., Azizi, S., Tu, T., Mahdavi, S.S., Wei, J., Chung, H.W., Scales,
  N., Tanwani, A., Cole-Lewis, H., Pfohl, S., et~al.: Large language models
  encode clinical knowledge. Nature  \textbf{620}(7972),  172--180 (2023).
  \doi{10.1038/s41586-023-06291-2}

\bibitem{tabuada2007event}
Tabuada, P.: Event-triggered real-time scheduling of stabilizing control tasks.
  IEEE Transactions on Automatic Control  \textbf{52}(9),  1680--1685 (2007).
  \doi{10.1109/TAC.2007.904277}

\bibitem{tartakovsky2014sequential}
Tartakovsky, A., Nikiforov, I., Basseville, M.: Sequential Analysis: Hypothesis
  Testing and Changepoint Detection. Monographs on Statistics and Applied
  Probability, Chapman and Hall/CRC (2014). \doi{10.1201/b17279}

\bibitem{wald1945sequential}
Wald, A.: Sequential tests of statistical hypotheses. Annals of Mathematical
  Statistics  \textbf{16}(2),  117--186 (1945). \doi{10.1214/aoms/1177731118}

\bibitem{zinkevich2003online}
Zinkevich, M.: Online convex programming and generalized infinitesimal gradient
  ascent. In: International Conference on Machine Learning (ICML). pp. 928--936
  (2003)

\end{thebibliography}
